\documentclass[letterpaper, 10 pt, conference]{ieeeconf}  
\usepackage{amssymb}
 \usepackage[pdftex]{graphicx}
 \usepackage{amsmath}
\usepackage{balance}
 \newcommand{\argmax}{\mathop{\rm arg~max}\limits}

\IEEEoverridecommandlockouts                              

\title{\LARGE \bf
Cooperative Perception with Deep Reinforcement Learning\\for Connected Vehicles
}

\author{Shunsuke Aoki$^{1}$, Takamasa Higuchi$^{2}$, and Onur Altintas$^{2}$
\thanks{$^{1}$Shunsuke Aoki is with Department of Electrical \& Computer Engineering, Carnegie Mellon University
        {\tt\small shunsuka@andrew.cmu.edu}}%
\thanks{$^{2}$Takamasa Higuchi and Onur Altintas are with InfoTech Labs, Toyota Motor North America R\&D
        {\tt\small \{takamasa.higuchi, onur.altintas\}@toyota.com}}%
}

\begin{document}

\maketitle
\thispagestyle{empty}
\pagestyle{empty}

\begin{abstract}

Sensor-based perception on vehicles are becoming prevalent and important to enhance the road safety.
Autonomous driving systems use cameras, LiDAR, and radar to detect surrounding objects, while human-driven vehicles use them to assist the driver.
However, the environmental perception by individual vehicles has the limitations on coverage and/or detection accuracy.
For example, a vehicle cannot detect objects occluded by other moving/static obstacles.
In this paper, we present a cooperative perception scheme with deep reinforcement learning to enhance the detection accuracy for the surrounding objects.
By using the deep reinforcement learning to select the data to transmit, our scheme mitigates the network load in vehicular communication networks and enhances the communication reliability.
To design, test, and verify the cooperative perception scheme, we develop a {\it Cooperative \& Intelligent Vehicle Simulation (CIVS) Platform}, which integrates three software components: traffic simulator, vehicle simulator, and object classifier.
We evaluate that our scheme decreases packet loss and thereby increases the detection accuracy by up to $12 \%$, compared to the baseline protocol.
\end{abstract}

\section{INTRODUCTION}

Cooperative perception is an emerging technology to enhance road safety by having connected vehicles exchange their raw or processed sensor data with the neighboring vehicles over Vehicle-to-Vehicle (V2V) communications \cite{kim2015impact}.
In fact, ETSI (European Telecommunications Standards Institute) has launched the standardization of cooperative or collective perception \cite{ETSI_CPM, ETSIApplication}.
For both human-driven and autonomous driving vehicles, sensor data capturing for the blind spot is helpful to avoid vehicle collisions and deadlocks.
The environmental perception by local on-board sensors has limitations on the coverage and/or detection accuracy.
When the target object is far away from the sensor or occluded by other road objects, it might not be detected and/or classified accurately.




While cooperative perception is a promising solution to enhance the sensing capabilities of connected vehicles, it heavily relies on V2V communications and generating non-negligible amount of data traffic.
In particular, when the road is congested with a number of connected vehicles, multiple vehicles may repeatedly send redundant information about the same object, wasting the network resources.
The excessive network load would increase the risk that important data packets are delayed or even lost, potentially leading to serious safety concerns.
To keep the communication reliability and road safety, each connected vehicle should intelligently select the data to transmit to save network resources.

\begin{figure}[!t]
\centering
\includegraphics[width=9.00cm]{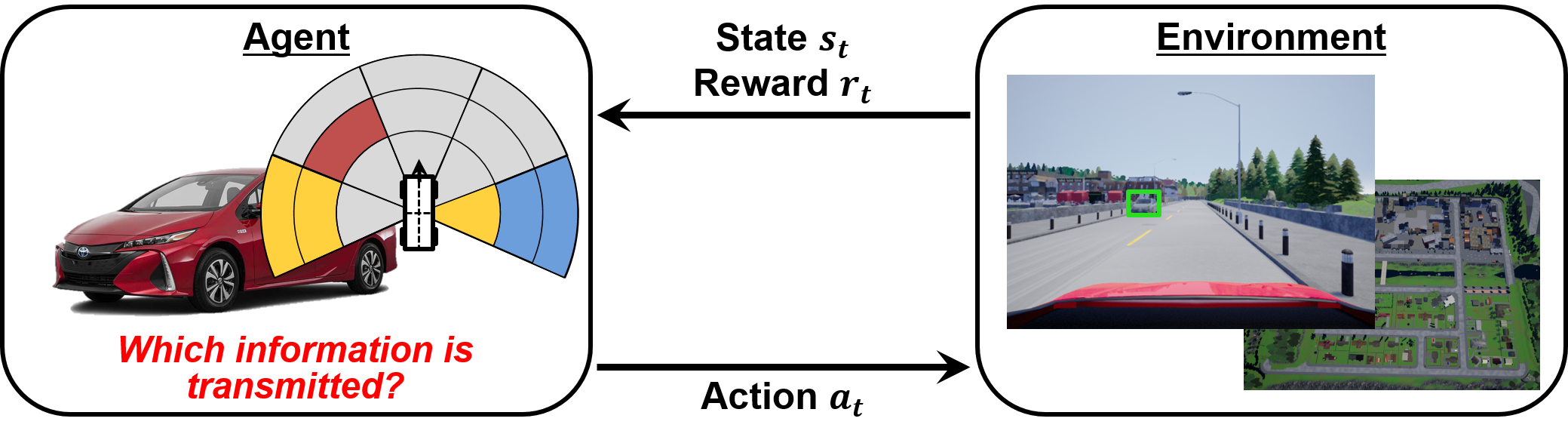}
\caption{Reinforcement Learning for Cooperative Perception.}
\label{fig:DeepRL}
\end{figure}

In this paper, we present a deep reinforcement learning approach for cooperative perception to mitigate the network load in vehicular communications.
We use the deep reinforcement learning for each connected vehicle to intelligently identify pieces of perception data worth to transmit, as shown in Figure \ref{fig:DeepRL}.
In our model, each connected vehicle uses the information from its local on-board sensors and vehicular communications to understand the current situation, and determines the packet transmission after processing the information by Convolutional Neural Networks (CNNs).
Our model mitigates the network load, avoids packet collisions, and finally enhances road safety and communication reliability.


It is not easy to test such cooperative perception systems using real vehicles because of cost and safety \cite{bhat2018tools, aoki2017merging}.
We design and develop a {\it Cooperative \& Intelligent Vehicle Simulation (CIVS) Platform}, where we integrate multiple software components to constitute a unified framework to evaluate a traffic model, vehicle model, communication model, and object classification model.
By using the CIVS platform, we can easily collect the vehicle mobility data and sensor data to train CNNs for the deep reinforcement learning.
The CIVS platform contains a vehicle simulator named CARLA \cite{dosovitskiy2017carla}, which provides realistic 3-D graphics and sensor models that can be used to test and verify the perception capabilities.
In addition, the CIVS platform uses the realistic vehicle mobility data that are generated by SUMO (Simulation of Urban MObility) traffic simulator \cite{behrisch2011sumo}.

The contributions of this paper are as follows.
\begin{enumerate}
  \item We present a deep reinforcement learning approach for cooperative perception to mitigate the network load.
  \item We design and develop a simulation environment with multiple open software and tools to assess our scheme in a repeatable manner.
  \item We evaluate our scheme with the simulation environment and demonstrate superior detection accuracy and reliability.
\end{enumerate}

The remainder of this paper is as follows.
Section I\hspace{-.1em}I describes the problem statement and overview of reinforcement learning.
Section I\hspace{-.1em}I\hspace{-.1em}I presents the deep reinforcement learning model we design and develop.
Section I\hspace{-.1em}V gives the simulation platform to test and verify our scheme.
Section V gives the simulation setup and the evaluation of our scheme.
Section V\hspace{-.1em}I discusses previous work related to our research.
Finally, Section V\hspace{-.1em}I\hspace{-.1em}I presents our conclusions and future work.

\begin{figure}[!t]
  \begin{center}
    \begin{tabular}{c}
      \begin{minipage}{0.48\hsize}
        \begin{center}
          \includegraphics[clip, width=3.5cm]{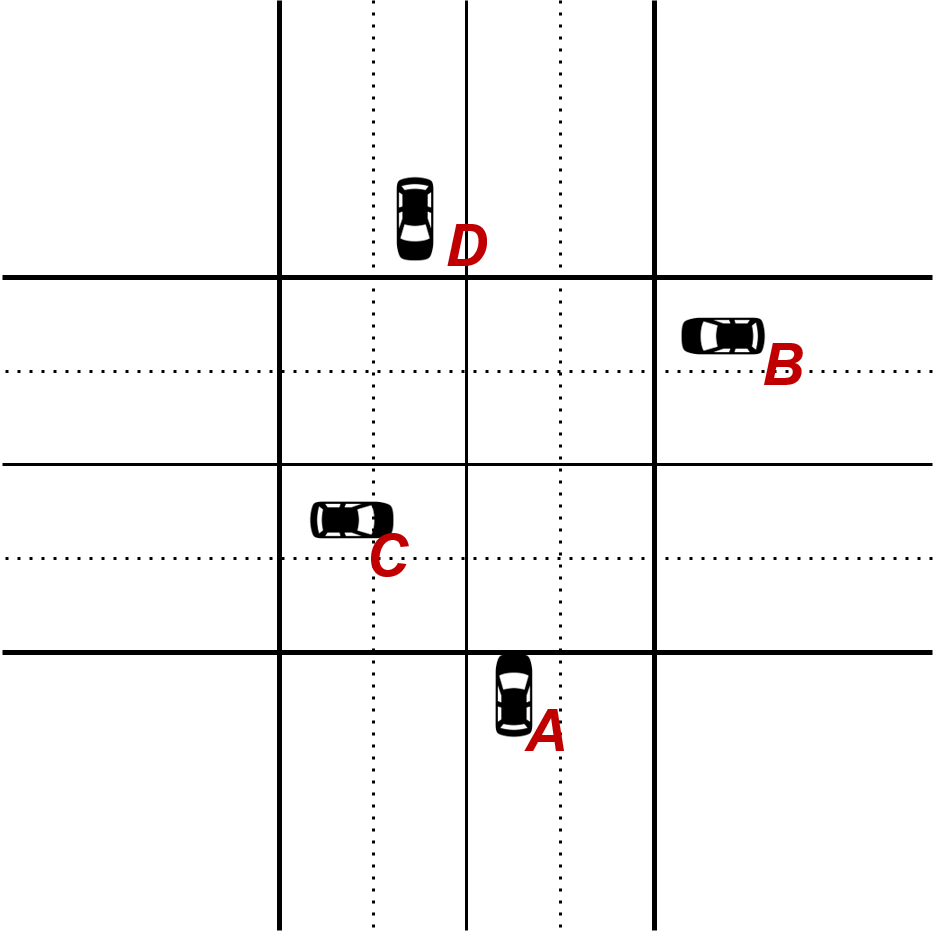}
	\hspace{1.6cm} (a) All vehicles can see\\each other.
        \label{fig:case1}
        \end{center}
      \end{minipage}
      \begin{minipage}{0.48\hsize}
        \begin{center}
          \includegraphics[clip, width=3.5cm]{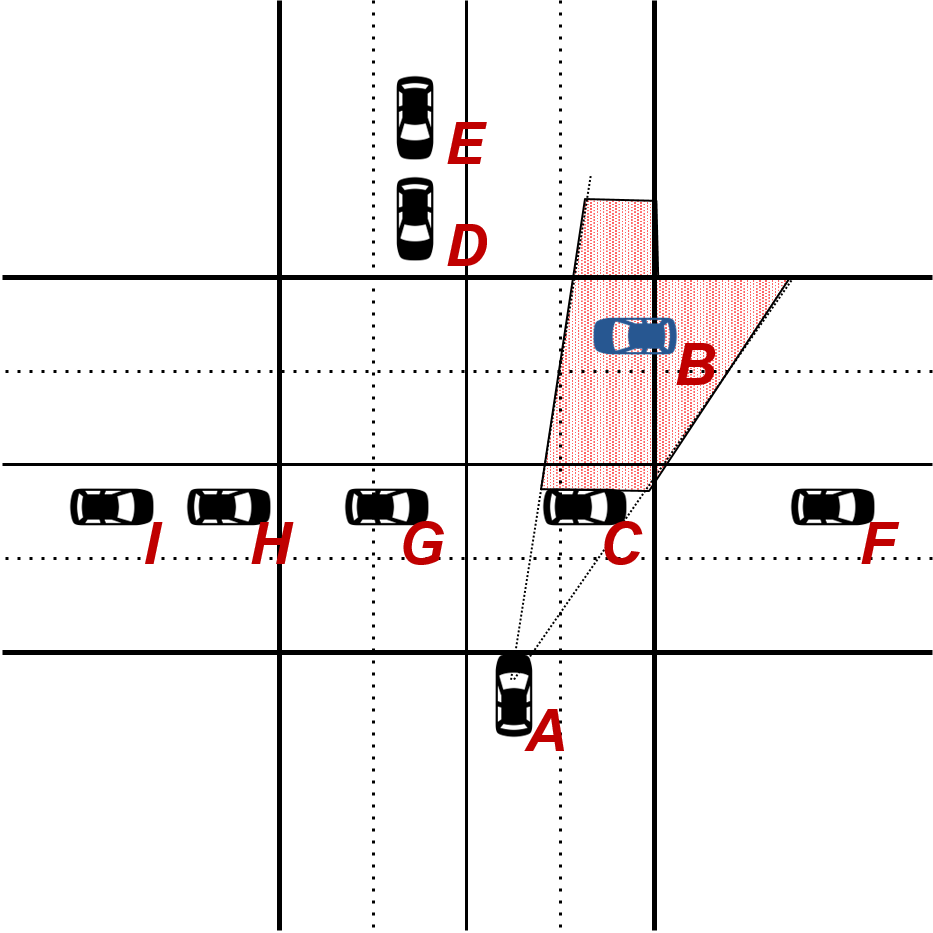}
	\hspace{1.6cm} (b) Cooperative Perception\\is required.
          \label{fig:case2}
        \end{center}
      \end{minipage}
    \end{tabular}
    \caption{Occlusion and Cooperative Perception.}
\label{fig:Case4CooperativePerception}
  \end{center}
\end{figure}

\section{PRELIMINARIES}

In this section, we present basic assumptions and formulate the problem on cooperative perception and reinforcement learning.

\subsection{Message Selection Problem in Cooperative Perception}

Cooperative perception enhances the perception capabilities of connected vehicles, but the vehicles need to intelligently select the data to transmit in order to save the network resources.
In this paper, we call this problem {\it Message Selection Problem in Cooperative Perception}.
In fact, to save the network resources and allocate the resources to the important packets, many researchers have studied the relevant problems \cite{higuchi2019cooperativeperception, gunther2016collective, allig2019dynamic}.

We present two example scenarios in Figure \ref{fig:Case4CooperativePerception}-(a) and -(b).
In these examples, the vehicles are equipped with on-board sensors (e.g., radar, LiDAR, and RGB camera) and a V2X communication interface.
They are within the communication range and are traversing the four-way intersection that is controlled by stop signs.
First, as shown in Figure \ref{fig:Case4CooperativePerception}-(a), when all the vehicles see each other, they have no need to share the perception data by the cooperative perception framework.
On the other hand, in Figure \ref{fig:Case4CooperativePerception}-(b), the vehicle A cannot see the vehicle B and cooperative perception messages from the surrounding vehicle(s) help drive safely.
At the same time, since the network resources are limited, when all the surrounding vehicles include all the perception data into their cooperative perception messages, some of the messages might be lost due to severe channel congestion. 
To keep the network reliability, it is desirable that connected vehicles in cooperative perception select the information that is likely to be beneficial to other vehicles in the vicinity.

\subsection{Reinforcement Learning}

Reinforcement learning is one of the popular machine learning techniques that enables an agent to learn its policy in an interactive environment, as shown in Figure \ref{fig:DeepRL}.
In this paper, we use the reinforcement learning to determine which pieces of information are transmitted by each connected vehicle, based on the surrounding context captured by local on-board sensors.

The reinforcement learning consists of three basic concepts: State, Action, and Reward.
The state describes the current situation of the agent.
The action is what the agent can do in each state.
Finally, the reward describes the positive or negative feedback from the environment by the action taken by the agent.
The overall goal in the reinforcement learning is learning a policy that maximizes the total reward.
Although there are many different techniques in reinforcement learning, we use Deep Q-Network \cite{mnih2015human} in this paper because of its simplicity and powerfulness. Deep Q-Network has two features: (i) Extension of Q-learning and (ii) Q-learning with Deep Neural Networks. Firstly, in Q-learning, we create and maintain a Q-table that is a reference table for the agent to select the best action. The agent can look up the Q-table to identify the rewards associated with all the state-action pairs. In the training period, we keep calculating and updating the Q-value stored in the Q-table as described in Eq. (\ref{eq:qlearning}).

\begin{equation}
Q^{new}(s_t, a_t) \leftarrow (1-\alpha)\cdot Q(s_t, a_t) + \alpha (r_t + \gamma \max_a Q(s_{t+1}, a))
\label{eq:qlearning}
\end{equation}

where $\alpha$ is the learning rate, $r_t$ is the reward, $\gamma$ is the discount factor, and $\max_a Q(s_{t+1}, a)$ is the estimated reward from the next action.
As described in Eq. (\ref{eq:actionspace}), the agent selects the optimal action to maximize the reward.

\begin{equation}
a_t = \argmax_{a \in A} Q(s_t, a)
\label{eq:actionspace}
\end{equation}

Secondly, since the size of Q-table may become huge due to the numbers of states and actions, we use Convolutional Neural Networks (CNNs), instead of the Q-table, in Deep Q-Networks. In the Deep Q-Network, the input is the state of the agent and the output is Q-values for all possible actions for the state.
The design for the state, action, and reward is discussed in Section \ref{section:ourdeepmodel}.


\section{COOPERATIVE PERCEPTION WITH DEEP REINFORCEMENT LEARNING}\label{section:ourdeepmodel}

In this section, we present our system model and the network model for cooperative perception with deep reinforcement learning.

\subsection{Our System Model}

\begin{figure}[!t]
\centering
\includegraphics[width=8.00cm]{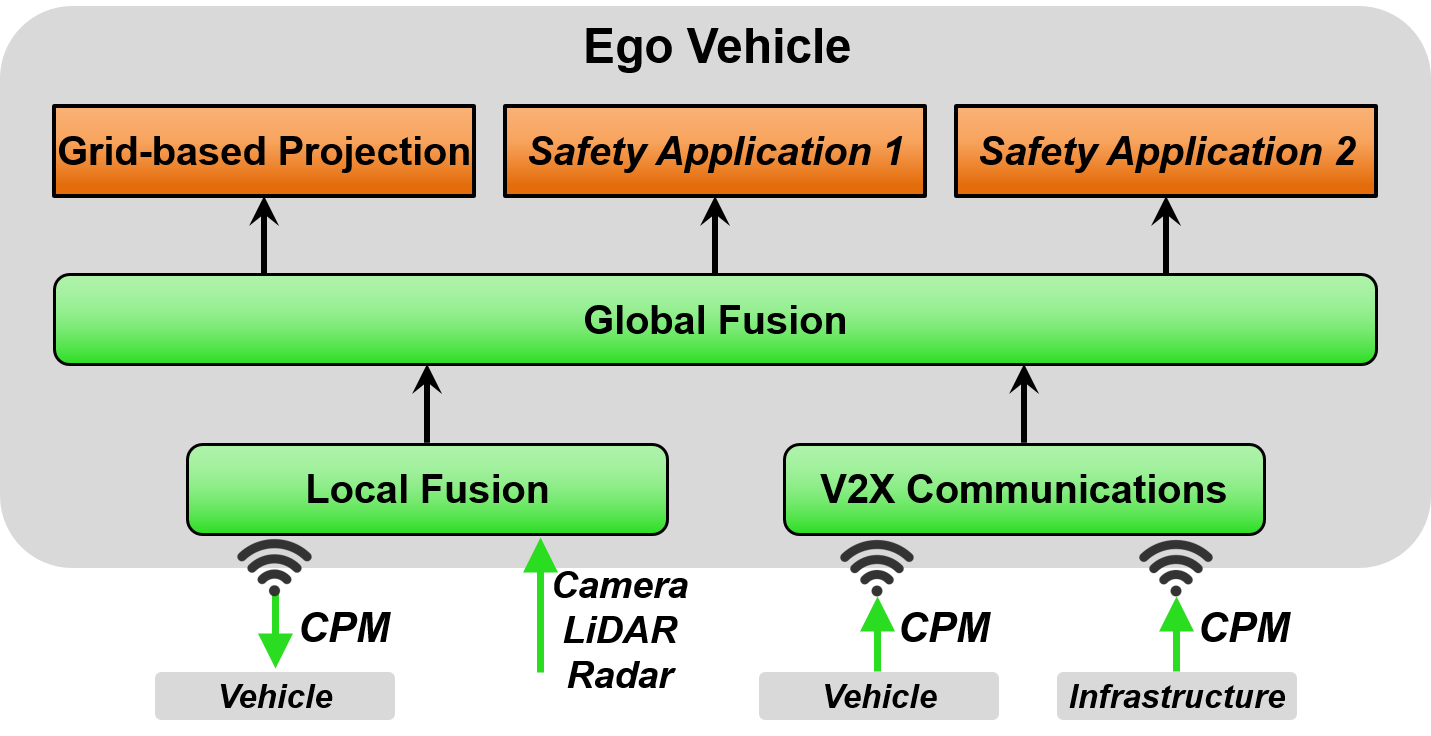}
\caption{Our Model for Sensor Fusion.}
\label{fig:SensorFusion}
\end{figure}

We first illustrate the architecture of our sensor fusion model in Figure \ref{fig:SensorFusion}.
In the model, each connected vehicle receives Cooperative Perception Message (CPM) via V2X communications from the neighboring connected vehicles and/or from roadside units.
In addition, the vehicles locally fuse the information from multiple on-board sensors, such as cameras, LiDARs, and radars.
After processing these two types of data, our model globally fuses the perception data recieved via V2X communication networks.
There are many different strategies \cite{rauch2011analysis} to fuse the information from local sensors and V2X communications.
Our model does not depend on any specific data fusion algorithms, while we prioritize the local perception information.
In addition, to avoid the information flooding and/or spreading rumors in the vehicular communications, our model only transmits the perception information based on the local on-board sensors.
After the global fusion, our model projects the information to the grid-based container \cite{birk2006merging} for the {\it State} $s_t$ in reinforcement learning.

From a standards viewpoint, each connected vehicle uses both Basic Safety Messages (BSM) and Cooperative Perception Messages (CPM) for V2V communications and they are broadcast at 10 Hz. BSM contains the position, speed, acceleration, and orientation of the ego vehicle. CPM contains the relative positions, orientation, and object type of the detected objects.

\begin{figure}[!t]
\centering
\includegraphics[width=3.75cm]{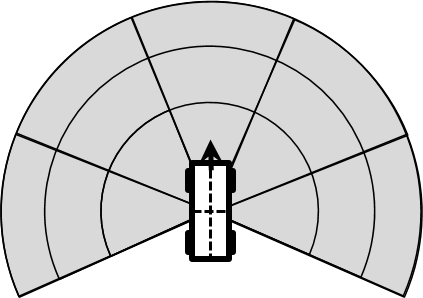}
\caption{Grid-based Circular Projection.}
\label{fig:circularprojection}
\end{figure}

\subsection{Deep Q-Learning and Network Model}\label{section:DeepReinforcement}

In this section, we present our deep reinforcement learning model and Convolutional Neural Networks (CNNs).
First, since the goal of reinforcement learning is to maximize the long-term rewards through the maneuvers, we design the states, actions, and rewards for cooperative perception as below:

\begin{table}[t]
\centering
\caption{Categories for Circular Projection.}
  \begin{tabular}{c|c|c|c|c} \hline
    Category & Local & BSM & CPM & Object \\ 
    ID & Perception & Transmission & Transmission & from CPM \\ \hline \hline 
    1 & Empty & x & x & x \\ \cline{2-5}
    2 & Occupied & x & x & x \\
    3 &  & x & x & \checkmark \\
    4 &  & \checkmark & x & x \\
    5 &  & \checkmark & x & \checkmark \\
    6 &  & \checkmark & \checkmark & x \\
    7 &  & \checkmark & \checkmark & \checkmark \\ \cline{2-5}
    8 & Occluded & x & x & x \\
    9 &  & x & x & \checkmark \\
    10 &  & \checkmark & x & x \\
    11 &  & \checkmark & x & \checkmark \\
    12 &  & \checkmark & \checkmark & x \\
    13 &  & \checkmark & \checkmark & \checkmark \\\hline
  \end{tabular}
\label{tb:13categories}
\end{table}

{\bf State:} We use two information for the state $s_t$: (i) Circular projection and (ii) Network congestion level. First, to maintain the perception data and history, we use the circular projection as shown in Figure \ref{fig:circularprojection}, where the part of the Field of View (FoV) is split into $5\times3$ grids.
Each grid has $1$ category from $13$ candidates, as shown in Table \ref{tb:13categories}.
The categories are determined by $4$ factors: Local perception, BSM (Basic Safety Messages) transmission, CPM (Cooperative Perception Messages) transmission, and Object reported from CPM.

As shown in Table \ref{tb:13categories}, the local perception classifies into $3$ categories: (i) {\it Empty}, (ii) {\it Occupied}, and (iii) {\it Occluded}.
Firstly, when there are no moving/static objects in the grid, the grid is labeled as Empty.
Secondly, when there is an object in the grid detected by the local on-board sensors, the grid is labeled as Occupied.
Also, when the grid is occluded by the object(s), the grid becomes Occluded in the projection.

As for the BSM, all the connected vehicles keep transmitting it as a safety beacon as specified in the standards.
For the CPM, since each agent controls transmission of the CPM based on its state, the agent may or may not receive the CPM from the neighboring connected vehicles.

In addition, we use the network load $\psi$ as part of the state $s_t$, because we have to select the data to transmit more strictly when the network is congested.
The network load $\psi$ is calculated from the number of BSMs and CPMs received during the recent time window.
In this paper, we represent the network load $\psi$ in $5$ levels.
When there are no surrounding vehicles, the network load $\psi$ becomes level $1$. On the other hand, when vehicle density is high as in a congested urban area, the value of $\psi$ becomes level $5$.
Although the agent cannot estimate the network congestion level for the receiver, since the agent and the receiver are supposed to be within the communication range of V2X communications, we assume they have the similar network conditions.
Overall, these information, including the circular projection and the network congestion, for the time window $W$ are used for the input for the CNNs.

{\bf Action:} The objective of our system is to save the network resources by decreasing the redundant messages in the vehicular communications while keeping the object tracking errors low. In our model, we define the action space $A = \{${\it Transmit}, {\it Discard} $\}$, where the agent broadcasts the CPM when the action becomes {\it Transmit} and the agent does not send the CPM when the action becomes {\it Discard}.
The action is calculated and determined by the Q-values that are the output of the CNNs.

{\bf Reward:} We design the reward for cooperative perception to decrease the duplicated information in the CPMs while enhancing the sensing capabilities.
We present our reward mechanism $r_{t, \omega, \alpha, \beta}$ in Eqs.(\ref{eq:rewardfunction}) and (\ref{eq:phiequation}), where we have $1$ reward and $3$ penalties.
$r_{t, \omega, \alpha, \beta}$ is the reward given at time $t$, for the target object $\omega$ in the communication from the transmitter $\alpha$ to the receiver $\beta$.

\begin{equation}
r_{t, \omega, \alpha, \beta} = \lambda_{local} + \mu_{CPM} \cdot \Theta_{t, \omega}+ \mu_{hist} \cdot \phi + \mu_{netcong} \cdot C_{t, \beta}
\label{eq:rewardfunction}
\end{equation}

\begin{equation}
\phi = \begin{cases}
    0 & (t-\tau_{\omega} > W) \\
    \frac{1}{(t- \tau_{\omega})} & (t-\tau_{\omega} \leq W)
  \end{cases}
\label{eq:phiequation}
\end{equation}

First, $\lambda_{local}$ is a binary reward, which becomes $1$ when the shared object $\omega$ is not detected by the receiver.
$\mu_{CPM}$, $\mu_{hist}$, and $\mu_{netcong}$ are the negative constant values, which are the penalties.
$\Theta_{t, \omega}$ represents the number of CPMs containing the information of the object $\omega$ at time $t$. By using this factor, our model can give the larger penalties when the multiple vehicles share the same information in the CPMs.
$\tau_{\omega}$ is the latest timestamp to detect the object $\omega$ by the local perception of the receiver $\beta$.
The objective of the cooperative perception is enhancing road safety by tracking the surrounding object(s). Thus, the vehicles do not need the information from the CPMs when the target object is detected very recently.
Finally, $C_{t, \beta}$ is the network congestion level at time $t$ for the receiver $\beta$.

\vspace{0.2cm}

For Deep Q-Learning, we design the CNNs that are composed of $3$ convolutional layers and $2$ fully connected layers. 
The first convolutional layer has $32$ kernels of $8\times8$ with stride of $2$, the second layer has $64$ kernels of $4\times4$ with strides of $2$, and the third convolutional layer has $64$ kernels of $3\times3$ with stride $1$. The fourth layer is a fully connected $512$ units and the last one has a unit for each action, Transmit and Discard.
To train and test the CNNs, as discussed in Section \ref{section:evaluation}, we collect the data from the simulation environment.


\begin{figure}[!t]
\centering
\includegraphics[width=7.90cm]{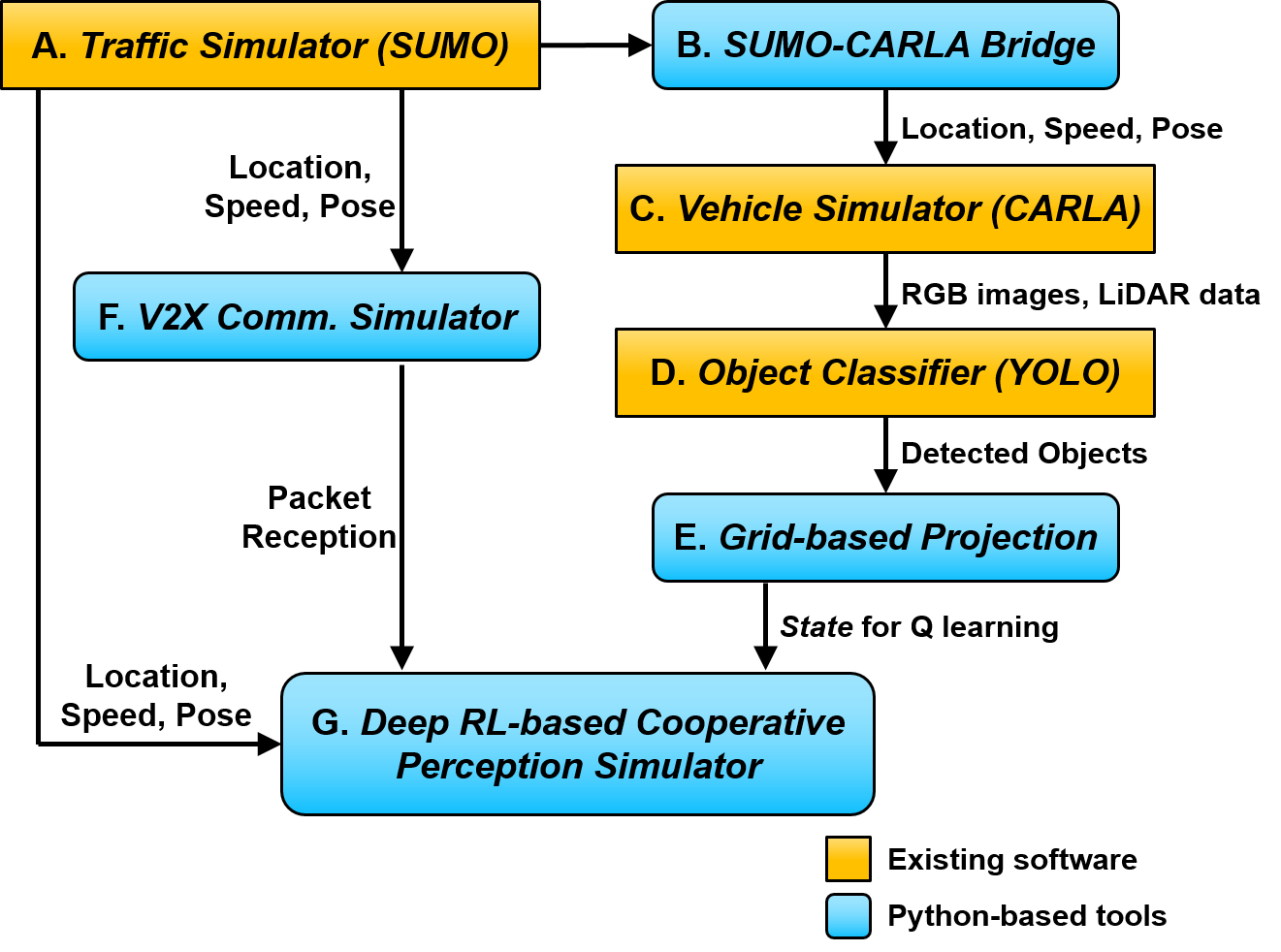}
\caption{Cooperative \& Intelligent Vehicle Simulation (CIVS) Platform.}
\label{fig:CIVSPlatform}
\end{figure}

\section{COOPERATIVE \& INTELLIGENT VEHICLE SIMULATION (CIVS) PLATFORM}

This section presents the Cooperative \& Intelligent Vehicle Simulation (CIVS) Platform that provides realistic 3-D graphics, traffic model, vehicle model, sensor model, and vehicular communication model.
The platform enables us to design, test, and verify the deep reinforcement learning approach for cooperative perception.
Figure \ref{fig:CIVSPlatform} presents the architecture of the CIVS platform, and it has seven components: (A) SUMO (Simulation of Urban MObility) traffic simulator \cite{behrisch2011sumo}, (B) SUMO-CARLA bridge, (C) CARLA vehicle simulator \cite{dosovitskiy2017carla}, (D) YOLO-based object classifier \cite{redmon2016you}, (E) Grid-based projection, (F) V2X communication simulator, and (G) Deep reinforcement learning-based cooperative perception simulator.

Since the CIVS platform consists of multiple components, we can easily test and verify the cooperative perception scheme under different settings. For example, when we test our applications under various traffic models, we only need to change the traffic simulator.

\subsection{SUMO Traffic Simulator}

SUMO \cite{behrisch2011sumo} is one of the most popular microscopic and open-source traffic simulator, which simulates realistic vehicular mobility traces.
In SUMO, each vehicle has its own route and moves individually through the road networks.
Also, we configure the Signal Phase and Timing (SPaT) for each traffic light.
In the CIVS platform, SUMO generates the mobility data for each vehicle, including the location, speed, and pose information.

\subsection{SUMO-CARLA Bridge: Trajectory Converter}

To keep the consistency between SUMO and CARLA, where the coordinate systems are different, we develop a Python-based tool named {\it SUMO-CARLA Bridge}.
The SUMO-CARLA bridge converts the mobility data from SUMO format into the CARLA format.
By using the mobility data from SUMO, the vehicles run in the CARLA world without any vehicle collisions and deadlocks and we can configure the traffic model to test and verify our applications\footnote{The source code is available at {\it https://github.com/BlueScottie/SUMO-Carla-integration}.}.

As for the road networks, SUMO simulator reads a set of XML files and CARLA uses OpenDRIVE file \cite{dupuis2010opendrive}.
The SUMO-CARLA bridge converts the road networks to keep the map consistency, as shown in Figure \ref{fig:ConsistentMap}.

\subsection{CARLA Vehicle Simulator}

CARLA \cite{dosovitskiy2017carla} is a game engine-based open-source simulator for automated vehicles and it supports flexible configurations of sensor suites and environmental conditions.
CARLA enables to test the sensor configuration and provides the realistic sensor data, such as RGB images and LiDAR data.
In addition, the software accommodates various types of vehicles with different colors and sizes. 
By using CARLA, the CIVS platform is able to test and verify the cooperative perception scheme with buildings and moving/static obstacles, which may occlude target objects.

\subsection{YOLO-based Object Classifier}

To process the sensor data to detect and classify the objects, the CIVS platform uses YOLOv3 \cite{redmon2016you}, which is one of the state-of-the-art object detection systems.
By using YOLOv3, we develop the list that contains the detected objects by each connected vehicle. In the detected objects list, each object has the information of the object type (car, pedestrian, traffic light etc.), the location (latitude and longitude), the distance from the sensor, and the relative direction from the sensor.

\subsection{Grid-based Projection}

After getting the objects list by the Object Classifier, our Python-based tool named {\it Grid-based Projection} maps the data to the grid-based sector.
The sector is split into $15$ grids as shown in Figure \ref{fig:circularprojection}, and each grid is classified as shown in Table \ref{tb:13categories}.
The projected sector is used to determine the state in the cooperative perception simulator.

\subsection{V2X Communication Simulator}

To simulate the vehicular communications model, we develop a Python-based tool named {\it V2X Communication Simulator}.
The V2X Communication Simulator gives a communication range and an interference distance for the vehicular communications.
The packet reception ratio depends on the network load.

\subsection{Deep Reinforcement Learning-based Cooperative Perception Simulator}

Finally, by using the vehicle mobility information, the vehicular networks information, and the grid-based projection, we run the Deep Q-Learning, which is discussed in Section \ref{section:DeepReinforcement}.
To train the convolutional neural networks, the platform may have to generate the sufficient amount of data with a variety of mobility patterns and vehicle patterns.




\begin{figure}[!t]
  \begin{center}
    \begin{tabular}{c}
      \begin{minipage}{0.49\hsize}
        \begin{center}
          \includegraphics[clip, width=3.4cm]{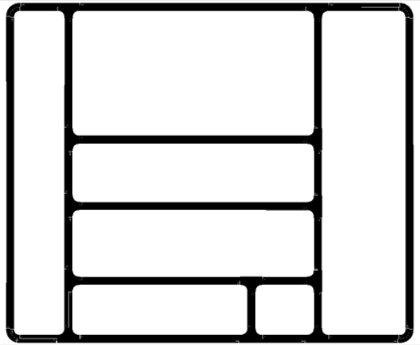}
	\hspace{1.6cm} (a) Road Network\\visualized in SUMO.
        \label{fig:sumoMap1}
        \end{center}
      \end{minipage}
      \begin{minipage}{0.49\hsize}
        \begin{center}
          \includegraphics[clip, width=3.8cm]{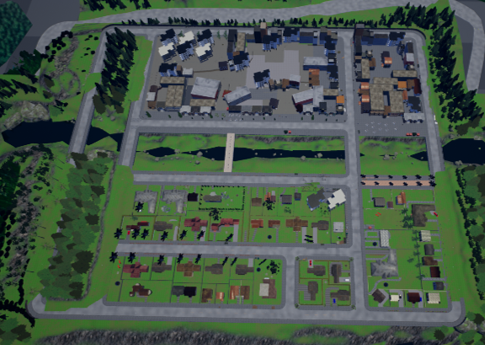}
	\hspace{1.6cm} (b) Road Network\\visualized in CARLA.
          \label{fig:carlaMap1}
        \end{center}
      \end{minipage}
    \end{tabular}
    \caption{Map Consistency in the CIVS Platform.}
\label{fig:ConsistentMap}
  \end{center}
\end{figure}

\section{EVALUATION} \label{section:evaluation}

We evaluate the cooperative perception scheme in terms of the network load and packet reception ratio, by using the CIVS platform.

\subsection{Simulation Scenario}

\begin{figure}[!t]
\centering
\includegraphics[width=6.70cm]{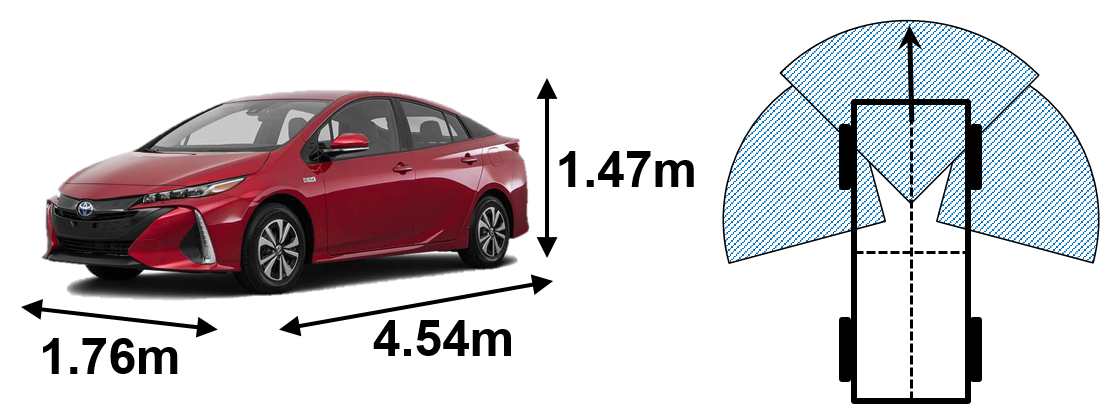}
\caption{Vehicle Size and Sensor Installation.}
\label{fig:camerainstallation}
\end{figure}

In the evaluation, we install $3$ RGB cameras on top of each connected vehicle to develop the circular projection, as shown in Figure \ref{fig:camerainstallation}.
Also, the connected vehicles have the same size: width is 1.76 meter, length is 4.54 meter, and height is 1.47 meter.
The scenario contains various types of vehicles and objects with different sizes, including buses, trucks, bicyclists, and pedestrians. Only the vehicles have network interfaces, and bicyclists are not connected.
In addition, the simulation world has the static objects, including trees, poles, traffic signs, and bus terminals that can block the sensor-based perception.

To train and test the CNN, we use two road networks that are provided by CARLA \cite{dosovitskiy2017carla} software: Map 1 and Map 2.
The data taken in Map 1 is used for training, and the data from Map 2 is used for testing. Figure \ref{fig:ConsistentMap} presents the Map 1 in SUMO and in CARLA.
To train the CNNs, we prepare $36$ scenarios, in which we have $4$ different vehicle densities: 50 (vehicles), 100 (vehicles), 150 (vehicles), and 200 (vehicles).
We have $9$ scenarios for each configuration for the vehicle density.
Each scenario has 160 seconds long. Overall, we record 200 hours ($=$160 sec$\times$(50+100+150+200)$\times$9) of sensor data in Map 1.
The vision cameras takes RGB image data at 10 Hz.

We evaluate our protocol by comparing against a baseline protocol. In the baseline protocol, each connected vehicle always broadcasts CPMs at 10 Hz whenever the perception detects moving/static objects.
Since the network resource is limited, data messages are not delivered when packet collisions occur.
At the same time, each connected vehicle broadcasts Basic Safety Messages (BSM) at 10 Hz.


\begin{figure}[!b]
\centering
\includegraphics[width=6.1cm]{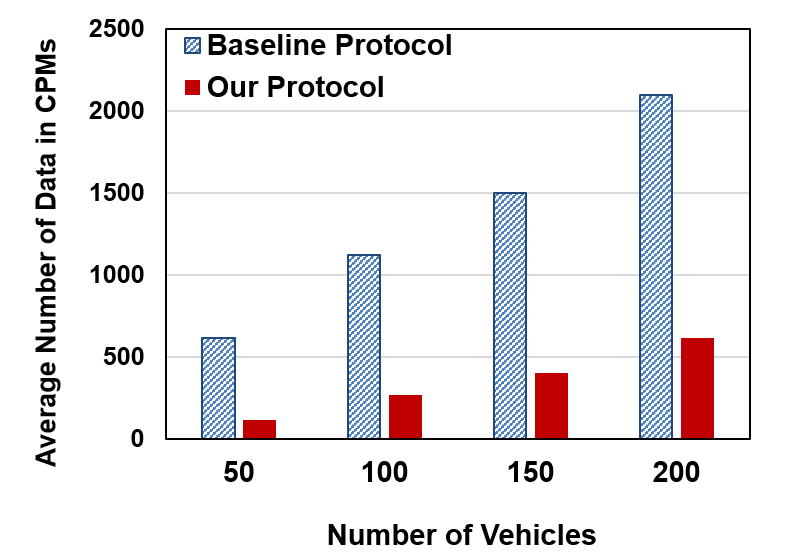}
\caption{Average Number of Data Shared in Cooperative Perception.}
\label{fig:totalnumberofCPMData}
\end{figure}

\subsection{Vehicular Communication Model}

For the vehicular communications, we set 300 meter as the communication range and 500 meter as the interference distance.
In our model, the CPM is successfully delivered to the neighbors with the probability $p$ \cite{higuchi2019cooperativeperception} as described below:

\begin{equation}
p = exp (- \lambda s / \gamma \tau)
\label{eq:packetreception}
\end{equation}

where $\lambda$ is the number of transmitters within the interference distance, $s$ is the average message size of CPMs, $\gamma$ is the data rate of vehicular communications, and $\tau$ is the transmission interval of CPMs.
In the experiments, we set $\gamma$ as 6 Mbps based on the typical data rate of DSRC (Dedicated Short-Range Communications) and ITS-G5 networks.
The $\tau$ is also determined as 100 ms following the standard.
The parameters $\lambda$ and $s$ dynamically change over time due to road traffic and the number of perception records, respectively.

\begin{figure}[!t]
\centering
\includegraphics[width=7.0cm]{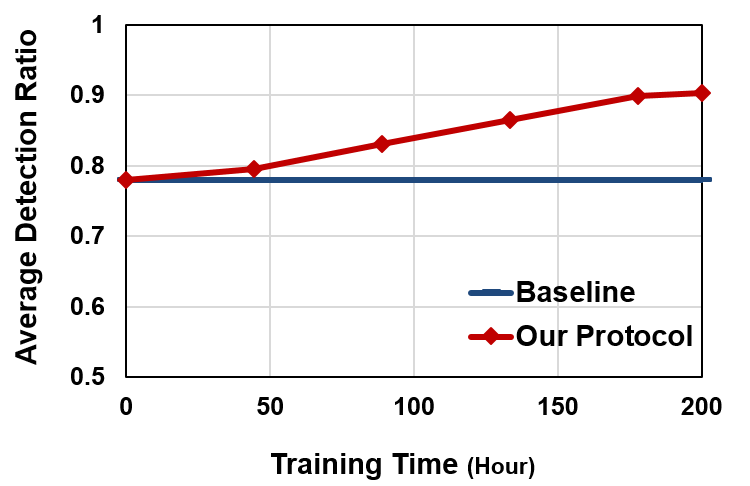}
\caption{Average Object Detection Ratio by Cooperative Perception.}
\label{fig:detectionratio}
\end{figure}

\begin{figure}[!t]
\centering
\includegraphics[width=7.0cm]{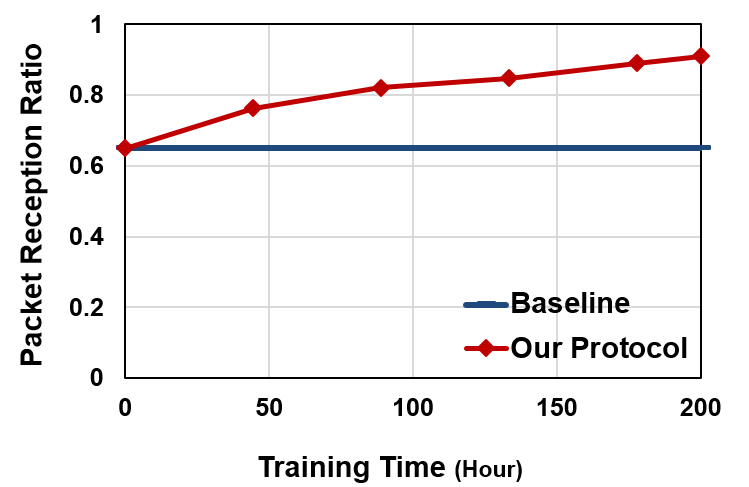}
\caption{Average Packet Reception Ratio in Cooperative Perception.}
\label{fig:packetdeliveryratio}
\end{figure}

\vspace{0.3cm}

\subsection{Simulation Results}

We run the simulations for evaluating the network load and object detection reliability of our deep reinforcement learning approach and compare against the baseline protocol.
First, we present the average number of data points shared in the CPMs for each vehicle in Figure \ref{fig:totalnumberofCPMData}.
We change the number of vehicles in the tests in Map 2, from 50 (vehicles) to 200 (vehicles), to understand the relationships between the vehicle density and the usage of our protocol.
As shown in Figure \ref{fig:totalnumberofCPMData}, our protocol hugely decreases the data shared in the CPMs regardless of vehicle density.

In addition, we present the average detection ratio by changing the training time in Figure \ref{fig:detectionratio}, to study the object detection reliability.
Note that the training is completed offline.
We simply use the vehicle mobility data from SUMO to get the ground truth, and measure the detection ratio within the range of $75$ (m) from each connected vehicle.
When the training time for the CNNs for Deep Q-Learning is insufficient, our protocol has the similar values on the detection probability as the baseline protocol, because of the network congestion and packet loss.
After training the CNNs, our protocol successfully increases the detection accuracy, at most $12 \%$, compared to the baseline protocol.

We also present the average packet reception ratio by changing the training time in Figure \ref{fig:packetdeliveryratio}.
Our protocol is always better than the baseline protocol, and the reception ratio is increased by $27 \%$ at most, because our protocol selects the data to share via the vehicular networks and the network load becomes light.

\vspace{0.3cm}

\section{RELATED WORKS}

Cooperative perception has been studied from the different aspects, such as sensor data processing, wireless networks, and its applications. First, many researchers have worked on the sensor fusion strategies \cite{rauch2011analysis, vasic2015collaborative, chen2019cooper} to increase the data reliability and consistency.
Rauch et al. \cite{rauch2011analysis} presented a two-step fusion architecture where processed sensor data are shared among neighboring vehicles. On the other hand, Chen et al. \cite{chen2019cooper} designed Cooper to enhance the detection ability for self-driving vehicles, where each vehicle shares the raw LiDAR data. Since the Cooper focused on the low-level fusion with raw sensor data, it spends much network resources on vehicular communications.

\vspace{0.12cm}

Second, to design effective and practical cooperative perception, there are multiple researches from vehicular communications \cite{higuchi2019cooperativeperception, gunther2016collective, gunther2015potential, ozbilgin2016evaluating}.
Gunther et al. \cite{gunther2015potential} studied the prospective impacts of the vehicular communications and collective perception by using Vehicular Ad-hoc Network (VANET) simulator called Veins. This research group also studied the feasibility of leveraging the standard decentralized congestion control mechanism \cite{ETSIAccessLayer}.
Higuchi et al. \cite{higuchi2019cooperativeperception} tackled the resource allocation for the vehicular communications by using the value-anticipating networks.


\vspace{0.12cm}

Third, there are many potential applications and usages of cooperative perception \cite{kim2015impact, aoki2019DSIP, kim2014multivehicle, aoki2018dynamic}. Kim et al. \cite{kim2015impact} investigated the impact of cooperative perception on decision making and planning of self-driving vehicles.
In addition, cooperative perception might enhance road safety in a variety of scenarios, such as road intersection management \cite{aoki2019DSIP} and overtaking/lane-changing maneuvers \cite{kim2014multivehicle}. To achieve the safe cooperation between self-driving vehicles and human-driven vehicles\cite{aoki2019DSIP, tsukada2019cooperative}, cooperative perception plays an important role because it enhances the perception capabilities of self-driving vehicles.


\vspace{0.12cm}

Deep reinforcement learning \cite{mnih2013playing} has become a popular technique even on vehicular communication.
For example, Ye et al. \cite{ye2018deep} utilized deep reinforcement learning for a decentralized resource allocation for the V2V communications. In this work, each V2V link finds optimal spectrum and transmission power.
Also, Atallah et al. \cite{atallah2017deep} present deep reinforcement learning model to design an energy-efficient scheduling policy for Road Side Units (RSUs) that meets both the safety requirements and Quality-of-Service (QoS) concerns.
When the optimization problem become complex, deep reinforcement learning might be a promising solution.
While these works studied the feasibility of deep reinforcement learning techniques for vehicular communications, there are no previous studies on reinforcement learning for cooperative perception.

%


\vspace{0.5cm}

\section{CONCLUSION}

\vspace{0.12cm}

We presented a cooperative perception scheme with deep reinforcement learning, where Connected Vehicles (CVs) intelligently select the data to transmit in order to keep the data traffic in vehicular networks low.
By decreasing the network load, the system can reduce the risk of packet collisions. As a result, our cooperative perception scheme enhances the detection accuracy and reliability.
We also presented the Cooperative \& Intelligent Vehicle Simulation (CIVS) Platform to design, test, and verify the cooperative perception scheme.
The CIVS platform provides realistic 3-D graphics, the traffic model, the vehicle model, the sensor model and the communication model to assess the feasibility and safety of our approach.




\vspace{0.12cm}

We finally note several limitations of our work. First, we need to consider the effects of buildings and objects on vehicular communications. Secondly, we used two road networks to evaluate our scheme in this paper. In future work, we will prepare additional road networks and evaluate our scheme with them.

\vspace{0.3cm}
\bibliographystyle{ieeetr}

\bibliography{IEEEabrv,bib_From20141215}

\end{document}